# Dependency length minimization:

## Puzzles and Promises

Haitao Liu[a,c], Chunshan Xu[a,b] and Junying Liang[a,1]

[a]Department of Linguistics, Zhejiang University, Hangzhou, CN-310058, China; [b]School of Foreign Languages, Anhui Jianzhu University, Hefei, CN-230601, China; and [c]Ningbo Institute of Technology, Zhejiang University, Ningbo, CN-315100, China.

**Abstract:**
In the recent issue of PNAS, Futrell et al. claims that their study of 37 languages gives the *first* large scale cross-language evidence for Dependency Length Minimization, which is an overstatement that ignores similar previous researches. In addition，this study seems to pay no attention to factors like the uniformity of genres, which weakens the validity of the argument that DLM is universal. Another problem is that this study sets the baseline random language as projective, which fails to truly uncover the difference between natural language and random language, since projectivity is an important feature of many natural languages. Finally, the paper contends an "apparent relationship between head finality and dependency length" despite the lack of an explicit statistical comparison, which renders this conclusion rather hasty and improper.

Key words:   dependency length minimization, cross-language, projectivity,

For decades, dependency length (distance) minimization has been pursued as a universal underlying force shaping human languages. In a recent issue of PNAS, Futrell, et al. (2015) suggest that dependency length minimization (DLM) is a universal property of human languages and hence supports explanations of linguistic variation in terms of general properties of human information processing. However, this statement is much exaggerated and far-fetched .

First of all, it is claimed in the paper that this is the *first* large scale cross-language evidence for DLM, since "previous comprehensive corpus-based studies of DLM cover seven languages in total". However, this is absolutely **NOT** true. In fact, there have been some large scale cross-language studies of DLM. For example, Liu (2008) has compared dependency distance of 20 natural languages with that of two different random languages, and pointed out that dependency distance minimization is probably universal in human languages. Evidently, the two articles share the

---

same research objective, the same research findings, and similar research methodologies.

There are some minor differences in the specific methods used in these two works. For example, Furtell et al (2015) hold dependency relations constant and draw random word order, while Liu (2008) held word order constant and drew random dependency relations. But such minor differences cannot deny the fact that the two works adopt similar research methods: both are based on the comparison between the dependency length (distance) of natural languages and that of corresponding artificial random languages. This method has also been used in an earlier study of two languages (Ferrer-i-Cancho 2004). The above difference in methods has no significant influence on the results of research, since it merely reflects the different ways to construct random languages in which the distribution of dependency length is randomized. Of course, it is perfectly acceptable and even encouraging for researchers to test previous findings with somewhat different methods. Anyway, any scientific finding must be subject to repeated tests. However, as far as this PNAS paper is concerned, we are much curious and puzzled why and how the authors could cite the work of Ferrer-i-Cancho and Liu (2014), which clearly introduces and largely dwells on previous DLM study based on 20 languages, but still claim that their PNAS paper is the *first* large scale cross-language evidence for DLM, and that "previous comprehensive corpus-based studies of DLM cover *seven* languages in total".

What is more, dependency length is sensitive to many factors. Linguistic properties, say DLM, may feature in one genre of language, but become vague and weak in another. Therefore, it is more desirable, especially in cross-language studies, to use a parallel corpus, or at least, corpora with the same genres, annotated with similar syntactic annotation schemes or drawn from native dependency treebanks (Jiang and Liu 2015). In the present study, however, it is not clear whether these conditions are satisfied by judging from the materials and methods, and hence there is some doubt in the validity of the argument that DLM is universal in all these languages.

As recently suggested, DLM bears closely on the rarity of crossing dependencies (Ferrer-i-Cancho 2013), and the authors also mention projectivity as one pervasive property of word order that can explain (or be explained by) DLM. What puzzles is that the baseline word order is set as projective. If projectivity is one feature of human language that contributes to DLM, it is desirable for a study of DLM to set baseline word order as non-projective so as to reveal the influence of projectivity on human languages in general. Projective baseline word order in this article fails to reveal the role of projectivity in DLM. In comparison, two baseline word orders respectively set as non-projective and projective may well throw much more light on DLM in natural languages, which has been adopted in previous works (Liu 2007, 2008). Also directly related to DLM is the distribution of dependency distance or the proportion of adjacent dependencies (AD) in natural languages. Previous studies have indicated that AD accounts for at least nearly half of all dependencies in any language investigated so far(Liu 2008), that the frequency of dependency drops dramatically with the increase of length (distance)(Liu 2007), and that a distribution of dependency distance is not influenced by variation in sentence length(Jiang and Liu 2015). These findings explain why DLM is persistently found in human languages and hence should have been mentioned in this article.

In the concluding part, the authors contend that an "apparent relationship between head finality and dependency length is a new and unexpected discovery". Nevertheless, it seems not apparent enough that dependency length is directly related to head-dependent order: no explicit statistical comparison is made in the present paper. Hence, the conclusion seems rather hasty, lacking solid supporting data. Theoretically, SVO order is in favor of DLM, as has been mathematically proven by Ferrer-i-Cancho (2015). But language is complex, constrained by multiple factors whose interactions may lead to no significant distance difference between VO and OV languages. In fact, existent corpus-based researches point to no definite relations between head placement and dependency distance. Gildea and Temperley (2010) find that German, as an OV language, has longer dependency distance than English, a VO language, but Hiranuma (1999) finds no difference between English and Japanese, which is an OV language, while Liu (2008) finds that Chinese, which is a VO language, has the longest mean dependency distance in all the languages that have been investigated. More importantly, another study (Liu and Xu 2012) that has quantitatively investigated 15 different languages clearly suggests no correlation between dependency distance and head placement. These findings indicate that, for complex systems like language, it is too casual to draw a relation between them based on one single study.

Taken together, Futrell et al. intend to address the dependency length minimization as a universal quantitative property of human languages. However they do overstate the significance of their study: it is definitely not the **first** large scale evidence of DLM, but a repetition of some previous works, though with slightly different methods. Further, they do not include adequate non-cognitive factors in mind. Finally, this paper is impaired by a lack of systematic review and references to related studies mentioned above in particular and dependency grammar in general (Hudson 2010), and due to this lack, it is legitimate to question the originality of this study because it is largely dissociated and disconnected from previous findings.

Futrell et al. have potentially displayed an intriguing domain for large-scale cross-linguistic research on dependency distance. However, the methodology itself is basically a repetitive effort of previous studies, and the data presented are not sufficient enough to support the conclusions made in this paper. This work uses more languages than previous studies——probably thanks to the fact that much more dependency treebanks are available today than in the past. However, simply using more languages in the study is insufficient to amend the drawbacks mentioned above.